\documentclass{styles/svproc}
\usepackage{cite}
\usepackage{amsmath,amssymb,amsfonts}
\usepackage{algorithm}
\usepackage{bm}
\usepackage{graphicx}
\usepackage{siunitx}
\usepackage{caption}
\usepackage{subcaption}
\usepackage{textcomp}
\usepackage{booktabs}
\usepackage{xcolor}
\usepackage{url}
\usepackage{hyperref}
\DeclareMathAlphabet{\pazocal}{OMS}{zplm}{m}{n}
\def\BibTeX{{\rm B\kern-.05em{\sc i\kern-.025em b}\kern-.08em
    T\kern-.1667em\lower.7ex\hbox{E}\kern-.125emX}}

\usepackage[noend]{algpseudocode}
\begin{document}
\mainmatter              
\title{Planning and Inverse Kinematics of Hyper-Redundant Manipulators with VO-FABRIK\thanks{Equal contribution. Contact: \email{cristian.morasso@studenti.univr.it}\\
This work has been supported by Hibot Corp.\\
The authors thank Prof. Joshua Vaughan (Univ. Louisiana at Lafayette) for his technical support, and Prof. Paolo Fiorini (Univ. Verona) for his mentoring.}}
\titlerunning{VO-FABRIK}  
%
\author{Cristian Morasso\inst{1}$^*$ \and Daniele Meli\inst{1}$^*$ \and Yann Divet\inst{2} \and Salvatore Sessa\inst{2} \and Alessandro Farinelli\inst{1}}
\authorrunning{Cristian Morasso et al.} 
%
%
\institute{University of Verona, Verona, Italy
\and
Hibot corp., Tokyo, Japan
}

\maketitle              

\begin{abstract}
Hyper-redundant Robotic Manipulators (HRMs) offer great dexterity and flexibility of operation, but solving Inverse Kinematics (IK) is challenging. In this work, we introduce VO-FABRIK, an algorithm combining Forward and Backward Reaching Inverse Kinematics (FABRIK) for repeatable deterministic IK computation, and an approach inspired from velocity obstacles to perform path planning under collision and joint limits constraints. We show preliminary results on an industrial HRM with 19 actuated joints. Our algorithm achieves good performance where a state-of-the-art IK solver fails.
\end{abstract}

\section{Introduction}
Hyper-redundant Robotic Manipulators (HRMs) guarantee high dexterity and are particularly suitable for operation in highly constrained environments, e.g., aerospace, search-and-rescue, and maintenance \cite{mu2022hyper}.
However, planning and control of HRMs in Cartesian space becomes extremely challenging as the number of Degrees of Freedom (DoFs) increases. In fact, Inverse Kinematics (IK) cannot be solved in closed form, but task-specific optimization of joint configuration within constraints as joint limits and collision avoidance is required \cite{siciliano2008springer}.

Standard optimization algorithms either do not guarantee fast convergence to the solution \cite{wang2010inverse}, or do not ensure repeatability \cite{deo1997minimum}, which is essential for reliability in safety-critical scenarios as smart industry.
Similarly, purely geometric approaches require assumptions on the specific robot and task at hand \cite{mu2018segmented}, hence are hardly generalizable.
When considering complex kinematic structures as HRMs, gradient-free methods
, e.g., genetic algorithms \cite{ruppel2018cost}, or deep neural networks 
\cite{kouabon2020learning}
are prominent solutions.
Such algorithms are however expensive in terms of training data and computational time. Furthermore, they do not provide any repeatability or safety guarantee \cite{corsi2021formal}.
The drawbacks of optimization-based approaches become ever more severe when solving the combination of IK and path planning problem 
\cite{zhong2021collision}
.

In this paper, we introduce VO-FABRIK, an algorithm combining Velocity Obstacles (VO) \cite{fiorini1998motion} and Forward and Backward Reaching Inverse Kinematics (FABRIK) \cite{aristidou2011fabrik}. FABRIK guarantees repeatable iterative fast IK computation. VO is integrated into FABRIK iterations and used for path planning at the end effector. Differently from alternatives as artificial potential fields \cite{ginesi2021dynamic}, VO prunes unsafe (colliding) velocities, thus guaranteeing the fast computation of collision-free configurations. 
Existing FABRIK applications in robotics consider few DoFs \cite{santos2020m} or do not guarantee joint limits and collision avoidance \cite{zhao2021collision}. 
Instead, we test VO-FABRIK on a (simulated) HRM with 19 rotational DoFs from Hibot Corporation (Float Arm), intended to navigate in cluttered hazardous environments and acquire detailed data for inspection and maintenance. We show the increased performance of VO-FABRIK against BioIK \cite{bioik}, a state-of-the-art IK solver based on evolutionary algorithms.

\section{Background}
We now introduce the basics of FABRIK and VO.

\subsection{Velocity Obstacles (VO)}
The VO algorithm computes the set of admissible velocities $\pazocal{V}_a$ for a robot $R$, given a target $G$ and $N$ obstacles defined by their radii (assuming spherical shapes for simplicity), current position and relative velocity to $R$.
At a given time step, the robot would reach its target with a velocity $\bm{v}_{pref}$ towards $G$. To avoid collisions, for each $i$-th obstacle a \emph{collision cone} $CC_i$ is computed as the set of robot velocities which will cause collision with $i$-th obstacle.
$\bm{v}_{pref}$ may then be modified, in order to fit in $\pazocal{V}_a = \pazocal{V} \setminus \bigcup_{i=1}^N CC_i$, where $\pazocal{V}$ is the full velocity space for the robot.

\subsection{FABRIK}
\begin{figure*}
    \centering
    \begin{subfigure}{0.24\textwidth}
    \includegraphics[width=\textwidth]{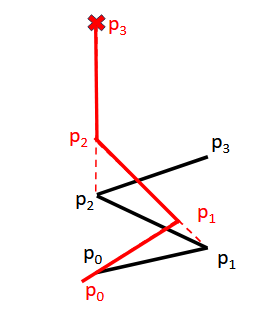}
    \caption{\protect\label{fig:fabrik}}
    \end{subfigure}
    \begin{subfigure}{0.24\textwidth}
    \includegraphics[width=\textwidth]{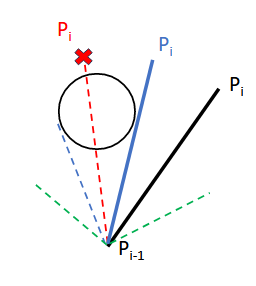}
    \caption{\protect \label{fig:vo_fabrik}}
    \end{subfigure}
    \begin{subfigure}{0.25\textwidth}
    \includegraphics[width=\textwidth]{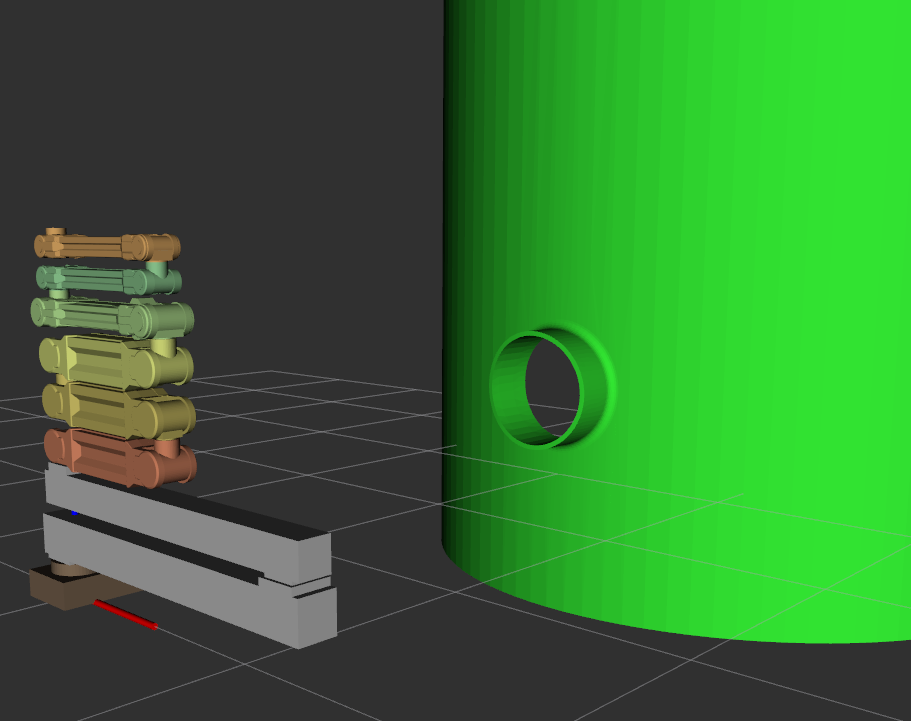}
    \caption{\protect\label{fig:setup}}
    \end{subfigure}
    \begin{subfigure}{0.225\textwidth}
    \includegraphics[width=\textwidth]{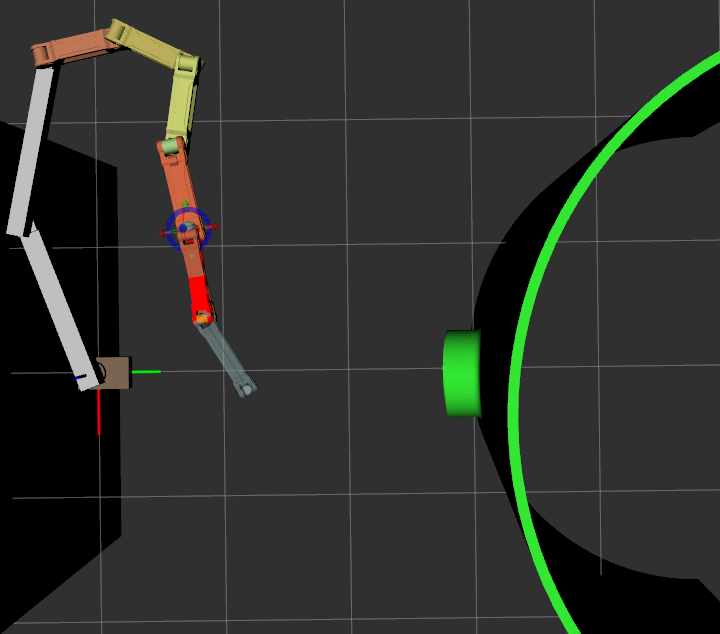}
    \caption{\protect\label{fig:setup_ext}}
    \end{subfigure}
    \caption{a) FABRIK backward iteration from black to red configuration; b) VO-FABRIK avoiding collision for link $i$, commanding blue configuration instead of red, avoiding the blue collision cone (obstacle) within the green cone (joint limits); the Float Arm c) in wrapped and d) in extended configuration.}
\end{figure*}
FABRIK algorithm solves the IK for a $N$-link kinematic chain iteratively, performing backward and forward phases and neglecting collision avoidance.
It starts from the goal end-effector position $\bm{p}_G$, and the positions $\{\bm{p}_i\}_{i=0}^N$ of link extrema ($\bm{p}_0$ being the base of the robot).
In the \emph{backward phase} (Figure \ref{fig:fabrik}), $\bm{p}_N$ is set to $\bm{p}_G$. Then, a segment from $\bm{p}_N$ to $\bm{p}_{N-1}$ is drawn, with the same length as $N$-th link. Its extremal point represents the new position $\bm{p'}_{N-1}$. The procedure is repeated for all links up to the base. Then, in the forward step, the reverse algorithm is applied, setting the newly computed $\bm{p'}_0$ to the original one (the base must remain fixed). Forward and backward steps are executed until $||\bm{p}_N - \bm{p}_G||_2 < \epsilon \in \mathbb{R}^+$. In this way, FABRIK guarantees minimal displacement between consecutive joint configurations.

\section{VO-FABRIK}
\begin{algorithm}
\caption{VO-FABRIK - IK Phase}\label{alg:Flow}
\begin{algorithmic}[1]
\Require positions of link extrema $\{\bm{p}_i\}_{i=0}^N$; length of links $\{l_i\}_{i=1}^N$; goal and base positions $\bm{p'}_N, \bm{p}_B$; set of obstacles $\pazocal{O}$; joint limits $\{jl_i\}_{i=0}^{N-1}$; joint angles $\{\alpha_i\}_{i=0}^{N-1}$
\While{not converged}
\State \textit{\% Backward phase}
\State $\bm{p}_N = \bm{p'}_N$
\For{$i=N-1; i \geq 0; i--$}
    \State $\bm{p}_i \gets $ \texttt{BACKWARD}($\bm{p}_{i+1}, \bm{p}_{i}, l_{i+1}$)
    \State $\pazocal{L} \gets $ \texttt{COMPUTE\_POS}($\{\bm{p}_J\}_{j=0}^i$) 
    \For{$o \in \pazocal{O} \cup \pazocal{L}$}
        \State Compute $CC$
        \State $jl_i \gets jl_i \cap $ \texttt{CONSTRAINTS}($CC, \bm{p}_{i+1}, \bm{p}_{i}$) 
    \EndFor
    \State $\alpha_i \gets $ \texttt{COMPUTE\_SAFE}($jl_i$)
    \State $\bm{p}_i \gets $ \texttt{FK}($\alpha_i$)
\EndFor
\State \textit{\% Forward phase}
\State $\bm{p}_0 = \bm{p}_B$
\For{$i=1; i \leq N; i++$}
    \State $\bm{p}_i \gets $ \texttt{FORWARD}($\bm{p}_{i+1}, \bm{p}_{i}, l_{i+1}$)
    \State $\ldots$ \textit{\% Analogous to backward phase, with inverted index order}
\EndFor
\EndWhile
\Return $\{\bm{p}_i\}_{i=0}^N, \{\alpha_i\}_{i=0}^{N-1}$
\end{algorithmic}
\end{algorithm}

VO-FABRIK consists of two phases: path planning and IK computation.
Given a target $G$ to be reached by the robot, in the \emph{motion planning phase} the velocity of the end effector is set to $\bm{v}_{pref}$ in the direction of $G$, with arbitrary module, resulting in a displacement from $\bm{p}_N$ to $\bm{p'}_N$ within a predefined time step $t_s$ (according to the truncated VO paradigm \cite{claes2012collision}). 
Then, in the \emph{IK phase} (Algorithm \ref{alg:Flow}), FABRIK starts with the backward step for each $i$-th link at Line 5.
Then, the closest point to $i$-th link is computed (Line 6) on each following link in the kinematic chain (hence, to the base). A virtual obstacle is added at these positions, and both virtual and real external obstacles are used to compute the set of collision cones $CC$ (Lines 7-10), with radius equal to the link's or obstacle's thickness. Since $i$-th link can only pivot around $\bm{p}_i$, collision cones for each obstacle here represent ranges of angular velocities (applied for $t_s$) which are safe for $i$-th link. An example is shown in Figure \ref{fig:vo_fabrik} in 2D, with blue lines representing $CC$ for a circular obstacle. $CC$ can then be converted to a range of feasible solid angles\footnote{Solid angles and joint limits can be converted to pitch and yaw by applying simple 2D projections, depending on the kinematic model of the robot.}, to be intersected with $i$-th joint limits (Line 11). Actual joint angles are then selected from the safe range (Line 12) and applied to update $\bm{p}_i$ with forward kinematics (Line 13). Similarly, the forward phase is performed and the process is iterated until convergence to $\bm{p'}_N$ is achieved.
The full VO-FABRIK algorithm terminates when $\bm{p}_G$ is finally reached.

\section{Experiments}
\begin{table}[t]
\centering
    \caption{Quantitative results.}
    \label{table:results}
    \begin{tabular}{c|c|c|c|c}
        \toprule
        & \multicolumn{2}{c|}{Wrapped (Fig. \ref{fig:setup})} & \multicolumn{2}{c}{Extended (Fig. \ref{fig:setup_ext})}\\
        & VO-FABRIK & BioIK& VO-FABRIK & BioIK\\
        \midrule
        Joint disp. [rad] & $0.01 \pm 0.13$ & N/A & $0.007 \pm 0.052$ & $0.004 \pm 0.025$\\
        Time per step [s] & $0.015 \pm 0.007$ & N/A & $0.015 \pm 0.007$ & $1.93 \pm 0.40$\\
        \bottomrule
    \end{tabular}
\end{table}

We implemented VO-FABRIK in Python, and tested it in a simulated environment with the Float Arm, shown in Figure \ref{fig:setup}. 
The Float Arm enters through a narrow opening and explores a cavity up and down.
The trajectory is calculated using a time step of $t_s =$ \SI{0.2}{s}.
Starting from home configuration shown in figure, BioIK\footnote{Collision checking is managed via MoveIt (\href{https://moveit.ros.org/}{https://moveit.ros.org/}), motion planning is based on VO. BioIK minimizes joint displacement.} is not able to complete the task, due to self-collisions. 
We then extend the home configuration as in Figure \ref{fig:setup_ext}.
Table \ref{table:results} shows that VO-FABRIK achieves comparable average joint displacement per step, while guaranteeing much faster computation per step.


\section{Conclusions and future works}
We presented VO-FABRIK, an algorithm for fast repeatable motion planning and IK guaranteeing safe collision avoidance within joint limits, thanks to the combination of FABRIK and VO. Our methodology is particularly suited for HRMs but can generalize to any kinematic configuration.
In the future, we will further assess the performance of our algorithm with multiple scenarios and robots. Moreover, we will extend the capabilities of our algorithm to also consider for link and end-effector orientation, currently not supported from FABRIK.

\bibliographystyle{ieeetr}
\bibliography{biblio}

\begin{thebibliography}{10}

\bibitem{mu2022hyper}
Z.~Mu {\em et~al.}, ``Hyper-redundant manipulators for operations in confined space: Typical applications, key technologies, and grand challenges,'' {\em IEEE Transactions on Aerospace and Electronic Systems}, 2022.

\bibitem{siciliano2008springer}
B.~Siciliano, O.~Khatib, and T.~Kr{\"o}ger, {\em Springer handbook of robotics}, vol.~200.
\newblock Springer, 2008.

\bibitem{wang2010inverse}
J.~Wang, Y.~Li, and X.~Zhao, ``Inverse kinematics and control of a 7-dof redundant manipulator based on the closed-loop algorithm,'' {\em International Journal of Advanced Robotic Systems}, vol.~7, no.~4, p.~37, 2010.

\bibitem{deo1997minimum}
A.~S. Deo and I.~D. Walker, ``Minimum effort inverse kinematics for redundant manipulators,'' {\em IEEE Transactions on Robotics and Automation}, vol.~13, no.~5, pp.~767--775, 1997.

\bibitem{mu2018segmented}
Z.~Mu, H.~Yuan, W.~Xu, T.~Liu, and B.~Liang, ``A segmented geometry method for kinematics and configuration planning of spatial hyper-redundant manipulators,'' {\em IEEE Transactions on Systems, Man, and Cybernetics: Systems}, vol.~50, no.~5, pp.~1746--1756, 2018.

\bibitem{ruppel2018cost}
P.~Ruppel, N.~Hendrich, S.~Starke, and J.~Zhang, ``Cost functions to specify full-body motion and multi-goal manipulation tasks,'' in {\em 2018 IEEE International Conference on Robotics and Automation (ICRA)}, pp.~3152--3159, IEEE, 2018.

\bibitem{kouabon2020learning}
A.~J. Kouabon, A.~Melingui, J.~M. Ahanda, O.~Lakhal, V.~Coelen, M.~Kom, and R.~Merzouki, ``A learning framework to inverse kinematics of high dof redundant manipulators,'' {\em Mechanism and Machine Theory}, vol.~153, p.~103978, 2020.

\bibitem{corsi2021formal}
D.~Corsi, E.~Marchesini, and A.~Farinelli, ``Formal verification of neural networks for safety-critical tasks in deep reinforcement learning,'' in {\em Uncertainty in Artificial Intelligence}, pp.~333--343, PMLR, 2021.

\bibitem{zhong2021collision}
J.~Zhong, T.~Wang, and L.~Cheng, ``Collision-free path planning for welding manipulator via hybrid algorithm of deep reinforcement learning and inverse kinematics,'' {\em Complex \& Intelligent Systems}, pp.~1--14, 2021.

\bibitem{fiorini1998motion}
P.~Fiorini and Z.~Shiller, ``Motion planning in dynamic environments using velocity obstacles,'' {\em The international journal of robotics research}, 1998.

\bibitem{aristidou2011fabrik}
A.~Aristidou and J.~Lasenby, ``Fabrik: A fast, iterative solver for the inverse kinematics problem,'' {\em Graphical Models}, vol.~73, no.~5, pp.~243--260, 2011.

\bibitem{ginesi2021dynamic}
M.~Ginesi, D.~Meli, A.~Roberti, N.~Sansonetto, and P.~Fiorini, ``Dynamic movement primitives: Volumetric obstacle avoidance using dynamic potential functions,'' {\em Journal of Intelligent \& Robotic Systems}, vol.~101, pp.~1--20, 2021.

\bibitem{santos2020m}
P.~C. Santos {\em et~al.}, ``M-fabrik: A new inverse kinematics approach to mobile manipulator robots based on fabrik,'' {\em IEEE Access}, 2020.

\bibitem{zhao2021collision}
L.~Zhao {\em et~al.}, ``Collision-free kinematics for hyper-redundant manipulators in dynamic scenes using optimal velocity obstacles,'' {\em International Journal of Advanced Robotic Systems}, 2021.

\bibitem{bioik}
P.~Ruppel {\em et~al.}, ``Cost functions to specify full-body motion and multi-goal manipulation tasks,'' in {\em 2018 IEEE International Conference on Robotics and Automation (ICRA)}, pp.~3152--3159, 2018.

\bibitem{claes2012collision}
D.~Claes, D.~Hennes, K.~Tuyls, and W.~Meeussen, ``Collision avoidance under bounded localization uncertainty,'' in {\em 2012 IEEE/RSJ International Conference on Intelligent Robots and Systems}, pp.~1192--1198, IEEE, 2012.

\end{thebibliography}

%
%







\end{document}